\documentclass[review,3p,12pt,times]{elsarticle}

\usepackage{amssymb}
\usepackage{amsmath}
\usepackage{graphicx}
\usepackage{xcolor}         
\usepackage{colortbl}
\usepackage{booktabs}

\journal{Pattern Recognition}

\begin{document}

\begin{frontmatter}



\title{FreeStyle: Free Lunch for Text-guided Style Transfer using Diffusion Models}


\author[label1]{Feihong He}
\ead{18996341802@163.com}

\author[label2]{Gang Li\corref{cor1}}
\ead{ucasligang@gmail.com}

\author[label3]{Fuhui Sun}
\ead{sunfh6732@163.com}

\author[label4]{Mengyuan Zhang}
\ead{maeyonzzzz@gmail.com}

\author[label2]{Lingyu Si}
\ead{lingyu@iscas.ac.cn}

\author[label3]{Xiaoyan Wang}
\ead{428163395@139.com}

\author[label1]{Li Shen}
\ead{mathshenli@gmail.com}

\cortext[cor1]{Corresponding author}

\affiliation[label1]{organization={School of Cyberspace Security at Sun Yat-sen University},
            city={Guangdong},
            postcode={518107},
            country={China}}

\affiliation[label2]{organization={Institute of Software,Chinese Academy of Sciences},
            city={Beijing},
            postcode={100190},
            country={China}}

\affiliation[label3]{organization={Information Technology Service Center of People's Court},
            city={Beijing},
            postcode={100745},
            country={China}}

\affiliation[label4]{organization={School of Computer Science and Technology, Harbin Institute of Technology},
            city={Shandong},
            postcode={264209},
            country={China}}

\begin{abstract}
The rapid development of generative diffusion models has significantly advanced the field of style transfer. However, most current style transfer methods based on diffusion models typically involve a slow iterative optimization process, e.g., model fine-tuning and textual inversion of style concept. In this paper, we introduce FreeStyle, an innovative style transfer method built upon a pre-trained large diffusion model, requiring no further optimization. Besides, our method enables style transfer only through a text description of the desired style, eliminating the necessity of style images. Specifically, we propose a dual-stream encoder and single-stream decoder architecture, replacing the conventional U-Net in diffusion models. In the dual-stream encoder, two distinct branches take the content image and style text prompt as inputs, achieving content and style decoupling. In the decoder, we further modulate features from the dual streams based on a given content image and the corresponding style text prompt for precise style transfer. Our experimental results demonstrate high-quality synthesis and fidelity of our method across various content images and style text prompts. Compared with state-of-the-art methods that require training, our FreeStyle approach notably reduces the computational burden by thousands of iterations, while achieving comparable or superior performance across multiple evaluation metrics including CLIP Aesthetic Score, CLIP Score, and Preference. We have released the code at: https://github.com/FreeStyleFreeLunch/FreeStyle.
\end{abstract}

\begin{graphicalabstract}
\end{graphicalabstract}

\begin{highlights}
\item A novel style transfer method (FreeStyle) based on diffusion models is proposed, leveraging pre-trained diffusion model parameters to achieve exceptional style transfer performance. This approach eliminates the need for style fine-tuning or inversion, significantly reducing computational costs.
\item The U-Net architecture within the diffusion model is explored, successfully decoupling image style and content. A dual-stream encoder is employed to separately encode style and content features, while a single-stream decoder merges these features multiple times during the upsampling process, achieving training-free transfer.
\item A feature modulation module is introduced, which scales and truncates content and style features in both the time and frequency domains, enabling precise control over the intensity of style and content representation. This provides flexibility in adjusting the strength of the transfer process.

\end{highlights}


\end{frontmatter}




\section{Introduction}
\begin{figure*}[t!]
\centering
\includegraphics[width=\textwidth]{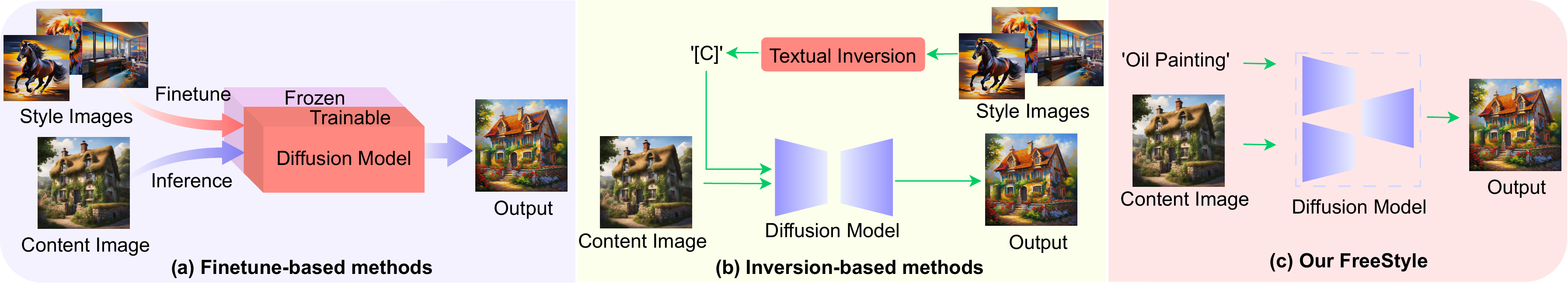}
\caption{Illustration depicting distinctions among fine-tune-based, inversion-based, and our FreeStyle approaches. (a) Fine-tuning the entire model or specific parameters embeds a visual style into the output domain of the diffusion model. (b) Embedding a specific style or content into a new pseudo-word (e.g., `[C]') via training set inversion, and using prompts with this pseudo-word to achieve style transfer. (c) Unlike the above methods, FreeStyle requires no optimization and utilizes the intrinsic style reconstruction ability of the diffusion model for effective style transfer.}
\label{problem2}
\end{figure*}

Image style transfer intends to transfer the natural image into the desired artistic image while preserving the content information. With the recent rapid development of generative diffusion models~\cite{stablediff,sdxl,PR1}, image style transfer has also witnessed significant advancements. These methods can be broadly classified into two categories: finetuning-based methods~\cite{stylediffusion} and inversion-based methods~\cite{InST,NullInversion}. The former (depicted in Fig.~\ref{problem2} (a)) requires optimizing some or all parameters to degrade the model to generate images of specific styles, while the latter (illustrated in Fig.~\ref{problem2} (b)) involves learning the specific style concept as the textual token to guide style-specific generation. Both approaches often require thousands or even more iterations of training, leading to significant computational costs and a slow optimization process.

Large text-guided diffusion models~\cite{stablediff}, on the other hand, are typically trained on large-scale datasets of text-image pairs, e.g., LAION dataset~\cite{laion}, which encompasses various style images and corresponding style text prompts. Consequently, these models~\cite{stablediff,sdxl} inherently possess the generative ability for specific styles. 
Recent works~\cite{alaluf2023cross} have introduced a cross-image attention mechanism to pre-trained diffusion models, enabling control of appearance or style transfer without optimization. However,  the use of appearance images or style images as references is still required. In some applications, users may not have access to reference images but want to engage in image transfer based on style text prompts. For instance, users can envision transforming their photos into styles reminiscent of Picasso or Da Vinci without possessing works by these renowned artists.

In this paper, we present a novel style transfer approach that requires neither optimization nor style images. Specifically, we propose a novel structure composed of a dual-stream encoder and a single-stream decoder. In this configuration, the dual-stream encoder separately encodes the content image and style text prompt as inputs, extracting features from the corresponding modalities for integration in the decoder. 
It has been demonstrated that the low-frequency signals and high-frequency signals of an image are strongly correlated with its semantic information and style information~\cite{seo2020dictionary}, respectively. We instantiate two modulation factors to balance low-frequency features from the U-Net's main backbone and high-frequency features from skip connections to implement image style transfer.
The first scaling factor regulates the strength of style transfer in the image and the second scaling factor controls the degree of content preservation in the images. Our approach is extremely simple and efficient, requiring only the adjustment of appropriate scaling factors to achieve the transfer of a specific style for any image.  

Through strategically modulating feature maps from U-Net’s skip connections and backbone, our FreeStyle framework exhibits seamless adaptability of style transfer when integrated with the existing large text-guided diffusion models, e.g., SDXL~\cite{sdxl}. It is important to note that despite structural differences from the U-Net~\cite{unet} model in pre-trained diffusion models, our approach incorporates U-Net modules without introducing new parameters. To our knowledge, FreeStyle is the first style transfer method based on diffusion models that neither requires reference style images nor any optimization. We conduct a comprehensive comparison of our method with other state-of-the-art techniques, including CLIPstyler~\cite{clipstyler}, CAST~\cite{CAST}, StyTr$^2$~\cite{stytr2}, UDT2I~\cite{uncoverdis}, etc. 

Our contributions are summarized as follows:
\begin{itemize}
\item We propose a simple and effective framework based on large text-guided diffusion models, called FreeStyle, which decouples the input of content image and textual input of desired style for specific style transfer without any optimization and style reference.
\item To further balance the preservation of content information and artistic consistency, we propose a novel feature fusion module designed to modulate the features of image content and the corresponding style text prompt.
\item We conduct comprehensive experiments with a wide range of content images and various style texts. The results show that the art images generated by FreeStyle exhibit accurate style expression and high-quality content-style fusion. Compared to state-of-the-art methods, FreeStyle demonstrates superior and more natural stylization effects. 
In our quantitative experiments, FreeStyle's CLIP Aesthetic Score improved by 1.4\% over others, Preference surpassed others by 32\%, and it also showed competitive performance in CLIP Score.
\end{itemize}

\section{Related Work}
\subsection{Image Style Transfer}
The field of style transfer plays a pivotal role in image processing and computer vision. It has seen a rapid evolution from manual texture synthesis to advanced neural style transfer (NST)~\cite{PR2,stytr2,CAST}, marking a significant shift from traditional techniques to modern deep learning approaches. 
Generative Adversarial Networks (GANs)~\cite{GAN}, with impressive image generation capabilities, have been rapidly applied to style transfer tasks~\cite{cyclegan}, further advancing the development of the field. 
With the recent rapid development of generative diffusion models~\cite{DDPM}, significant progress has been made in image style transfer. 
These techniques can be classified into two main categories: finetune-based methods and inversion-based methods. 
Finetune-based methods~\cite{stylediffusion} optimize some or all of the model parameters using extensive style images, embedding their visual style into the model's output domain. In contrast, inversion-based methods~\cite{InST,NullInversion} embed style or content concepts into special word embeddings using style or content images and achieve style transfer with prompts containing these word embeddings. 
The aforementioned methods based on diffusion models require style images for training models, resulting in a slow optimization process. Recent works~\cite{alaluf2023cross} introduce a cross-image attention mechanism and develop a style transfer method that does not require any optimization. However, these methods still rely on style images as references. 
As a text-guided style transfer method, FreeStyle differs by modulating features of the diffusion model, leveraging its inherent decoupling ability for style transformation without the need for extra optimization or style reference images. 

\subsection{Text-guided Synthesis}
GAN-CLS~\cite{t2iGAN} is the first to achieve text-guided image synthesis of flowers and birds using recurrent neural networks~\cite{RNN} and Generative Adversarial Networks~\cite{GAN}. 
Subsequently, numerous efforts in text-guided image generation~\cite{PR3} have propelled rapid development in this field. 
Benefiting from the introduction of CLIP~\cite{CLIP}, the remarkable generative capabilities of text-to-image models~\cite{PR5,imagen} have garnered significant attention from researchers, driven by the advancements in diffusion models. 
In addition to generating images that match text descriptions, text-guided techniques are now widely used in various tasks such as image editing~\cite{uncoverdis,pr4}, image restoration~\cite{textIR}, and video synthesis~\cite{videodiff} etc. 
Tsu-Jui Fu et al.~\cite{textst0} argue that traditional style transfer methods, which depend on pre-prepared specific style images, are both inconvenient and creativity-limiting in practical applications. 
Following this, a new style transfer method that is guided by textual descriptions~\cite{clipstyler} has been introduced, offering enhanced flexibility and convenience. 
This not only simplifies complex artistic creation but also makes advanced image manipulation accessible to a broader audience without the need for specialized graphic design skills. As a result, text-guided image processing is revolutionizing the way we interact with and create visual content. 

\subsection{Deep Model Fusion}
Deep model fusion~\cite{li2023deep} endeavors to integrate multiple deep neural networks (DNNs) into a singular network, maintaining their inherent capabilities and even surpassing the performance of multi-task training~\cite{ainsworth2022git}. With the emergence of new large language models (LLMs), such as GPT-3~\cite{brown2020language}, GPT-4~\cite{achiam2023gpt}, T5~\cite{raffel2020exploring} and BERT~\cite{devlin2018bert}, there is increasing attention on applying weighted averaging (WA) techniques~\cite{lv2023parameter} to these models. For instance, B-tuning~\cite{you2022ranking} utilizes Bayesian learning to calculate posterior prediction distributions, thereby fine-tuning the top-K ranked pre-trained models based on their transferability. Zoo-tuning~\cite{shu2021zoo} aggregates the weights of pre-trained models with aligned channels to create a final model adapted to downstream tasks, addressing the high costs associated with migrating large models. For diffusion models, FreeU~\cite{freeu} strategically reweights the contributions of feature maps from U-Net's skip connections and backbone to effectively enhance the quality of the generated images without any training. In FreeStyle, we fuse two latent space embeddings from different modality inputs and decode the latent space representation, which has absorbed information from both inputs, to generate an image that integrates both style and content information. 


\section{FreeStyle}
\subsection{Preliminaries}
Diffusion models~\cite{DDPM} involve a forward diffusion process and a reverse denoising process. 
During the forward process, Gaussian noise $\epsilon$ is progressively added to the clean sample $x_0$, with the intensity of the added noise $\epsilon$ increasing as $t\in\left[1, 2, \dots ,T\right]$ increases. 
The noised image at step $t$ is obtained through the diffusion process: 
\begin{equation}
x_t=\sqrt{\bar{\alpha}_t}x_0+\sqrt{1-\bar{\alpha}_t}\epsilon, 
\end{equation}
where $\epsilon \sim \mathcal{N}\left(0,\mathcal{I}\right)$, $\alpha_t=1-\beta_t$, 
 $\bar{\alpha_t}=\prod\limits_{i=0}^t\alpha_i$, and $\beta_t \in \left(0,1\right)$ is a fixed variance schedule. 
Conversely, in the denoising process, $x_T$ is gradually transformed into the clean image $x_0$ by progressively predicting and removing the noise. The sampling from step $t$ to step $t-1$ can be represented as: 

\begin{equation}
x_{t-1}= \sqrt{\bar{\alpha}_{t-1}} \hat{x}_{0, t}\left(x_{t}\right) + \sqrt{1-\bar{\alpha}_{t-1}} \epsilon_{\theta}\left(x_{t}, t, p\right). 
\end{equation}

Here, $p$ represents the condition input (e.g., text prompt), and $\epsilon_{\theta}$ denotes the noise prediction network. 

\begin{figure*}[!t]
\centering
\includegraphics[width=\textwidth]{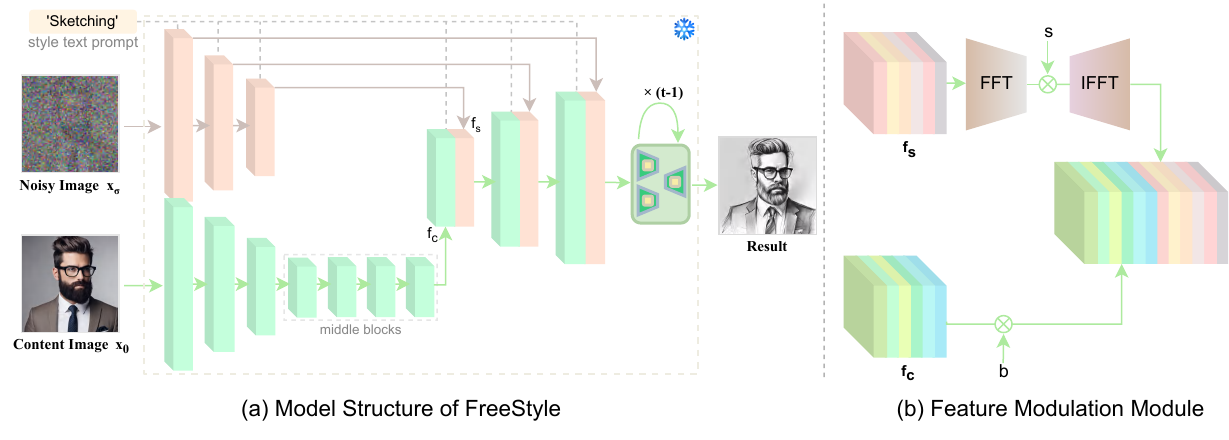}
\caption{\textbf{The overview of our FreeStyle Framework.} (a) \textbf{Model Structure of FreeStyle.} Our dual-stream encoder generates the content feature $f_c$ guided by the input content image $x_0$, and the style feature $f_s$ guided by the input style text prompt and noisy image $x_{\sigma}$. In the single-stream decoder, we modulate the content and style features through the feature modulation module. (b) \textbf{Feature Modulation Module.} Our feature modulation module refines style features $f_s$ and content features $f_c$ separately to ensure accurate style expression and complete content preservation. }
\label{overview}
\end{figure*} 

\subsection{Model Structure of FreeStyle}
In diffusion models, the U-Net structure is commonly used as the noise prediction network. It consists of an encoder and a decoder, along with skip connections that facilitate information exchange between corresponding layers of the encoder and decoder. 
We propose a novel modulation method for fusing content and style information in style transfer by balancing the low-frequency and high-frequency features from the U-Net's backbone and skip layers. 
Fig.~\ref{overview} (a) illustrates the overall structure of FreeStyle, which consists of a dual-stream encoder and a single-stream decoder. The dual-stream encoder in FreeStyle comprises two U-Net encoders with shared parameters, while the single-stream decoder is made up of the U-Net decoder structure. The dual-stream downsampling process can be described as follows:
\begin{equation}
\begin{cases}
f_s=&\mathrm{E}\left(x_{\sigma},p\right)\\
f_c=&\mathrm{E}\left(x_{0}\right), \\
\end{cases}
\end{equation}
where $p$ represents the embedding of the style text prompt, and $x_\sigma$ denotes the content image after $\sigma$ steps of noise addition. The $f_s$ and $f_c$ represent image features that carry style and content information, respectively. 
In the denoising process, we predict the noise distribution at step $t$ using the following formula:

\begin{equation}
    \epsilon_{t} \sim \mathcal{N}\left(\mu_{\theta'}\left(f_c,f_s,t\right),\Sigma_{\theta'}\left(f_c,f_s,t\right)\right),
\end{equation} 

where $\theta'$ represents the parameters of the decoder in the U-Net, $\mu_{\theta'}$ denotes the mean of the predicted noise distribution, and $\Sigma_{\theta'}$ indicates the variance of the distribution. Subsequently, we obtain the denoised image $x_{t-1}$ as follows:
\begin{equation}
    x_{t-1} = \sqrt{\bar{\alpha}_{t-1}} \hat{x}_{0, t}\left(x_{t}\right) + \sqrt{1-\bar{\alpha}_{t-1}} \epsilon_t, 
\end{equation}
where $\hat{x}_{0, t}\left(x_{t}\right)$ represents the estimate of $x_0$ given $x_t$ and $t$. 

\begin{figure*}[!t]
\centering
\includegraphics[width=\textwidth]{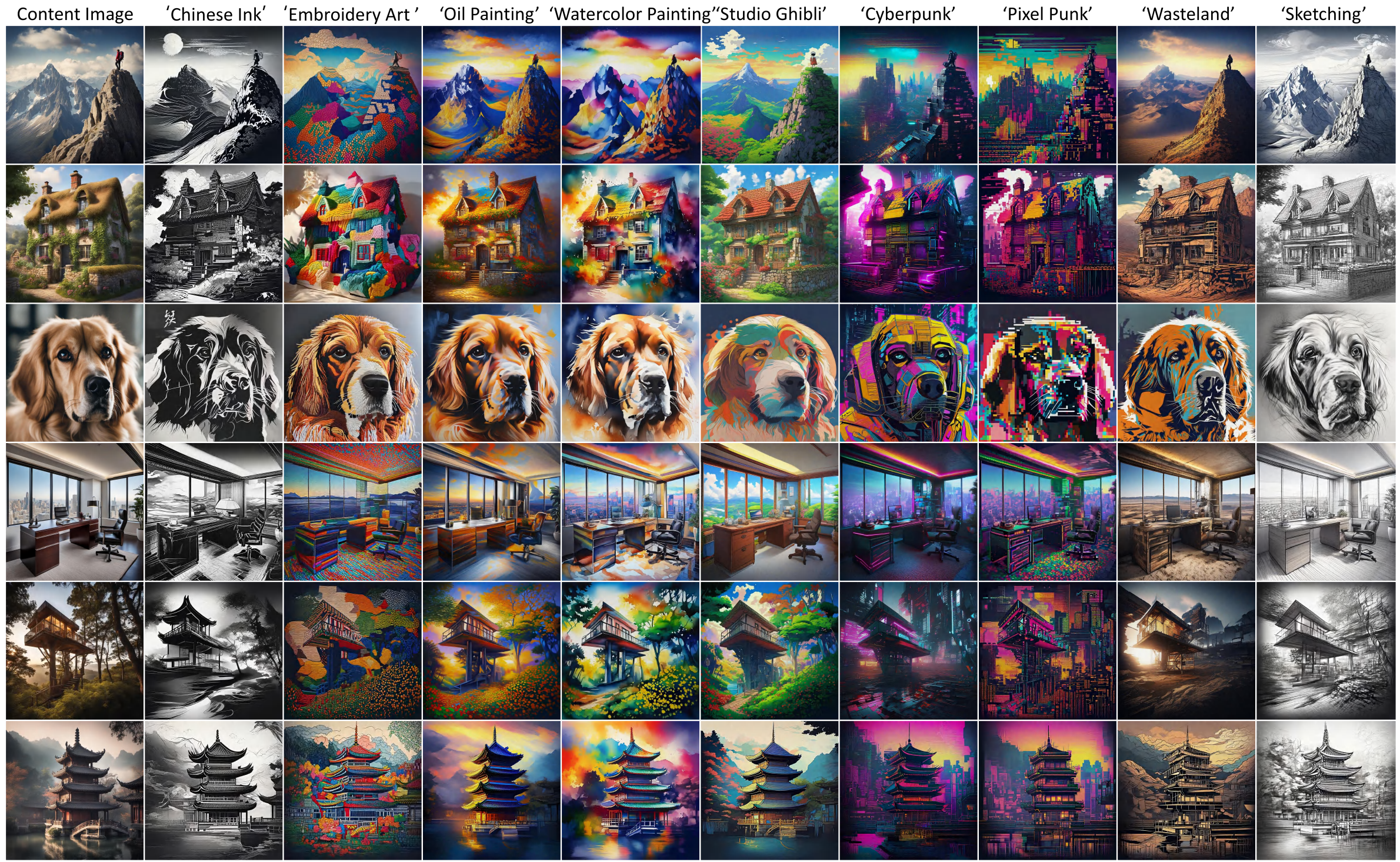}
\caption{Style transfer results using FreeStyle. Under training-free condition, our method can accurately express its style in images of various categories under various style text prompts, and can achieve a natural fusion of style and content.}
\label{result1}
\end{figure*}

\subsection{Feature Modulation Module}

We strategically reweight the contributions of feature maps from the encoders of two parameter-sharing U-Nets, effectively leveraging the strengths of both components to implement image style transfer. 
It has been demonstrated that images consist of low-frequency signals controlling image content and high-frequency signals governing image style~\cite{seo2020dictionary}. 
We implement an effective training-free style transfer by modulating the style feature $f_s$ and the content feature $f_c$ to complete artistic image generation. 

As shown in Fig.~\ref{overview}(b), the content feature $f_c$ is generated guided by the noise-free content image $x_0$, while the style feature $f_s$ is generated guided by the style text prompt $p$ and the noise-added image $x_{\sigma}$. 
During the upsampling process in U-Net, the features $f_c$ primarily influence the semantic expression of the generated result, while the features $f_s$ have a greater impact on the high-frequency detail information of the result. 
Consequently, we engage in special modulation of $f_s$ and $f_c$ to further activate the intrinsic style reconstruction capability of U-Net. 
To enhance the semantic characteristics of the feature $f_c$, we amplify their variance. Specifically, we apply a weight parameter $b$ (where $b$ \textgreater 1) to certain channels of the feature to expand their variance, the process can be represent as:
\begin{equation}
f_{c}'=concat\left(b\times f_{c}\left[:n\right], f_{c}\left[n:\right]\right),
\end{equation} 

where $n$ is the number of truncated channels of the feature, and $f_{c}'$ is the enhanced feature.
To suppress the low-frequency semantic characteristics while preserving high-frequency details and other style expression information, we first transform the feature $f_s$ into frequency domain information using the Fourier transform, and then apply a threshold $r_{\text{thresh}} = 1$ to filter out the low-frequency semantic information from the features. 
Subsequently, we use a weight parameter $s$ greater than 1 to enhance the style information. 
Finally, we convert the processed frequency domain features back into spatial domain features using the inverse Fourier transform. The process can be simply denoted as: 
\begin{equation}
f_{s}'=IFFT\left(\mathcal{F}\left(FFT\left( f_{s} \right)\right)\right),
\end{equation}
$FFT$ and $IFFT$ represent the Fourier transform and inverse Fourier transform, respectively. The function $\mathcal{F}$ is :
\begin{equation}
\mathcal{F}(r)=\left\{\begin{array}{ll}
s & \text { if } r<r_{\text {thresh }} \\
1 & \text { otherwise }
\end{array}\right.
\label{equa8}
\end{equation}
where $r$ is the radius. 
By applying the above operations, we modulate $f_c$ and $f_s$ to obtain $f_c'$ and $f_s'$, and finally concatenate $f_c'$ and $f_s'$ to feed them into the blocks of the U-Net decoder.

\section{Experiments}
In this section, we conduct extensive experiments on images from various domains such as buildings, landscapes, animals, and portrait. 
By performing qualitative and quantitative comparisons with the state-of-the-art style transfer methods, we validated the robustness and effectiveness of our approach. 
\subsection{Implementation Details}
Since our method is training-free, our method requires no training. Our experiments are conducted on an NVIDIA A100 GPU, with an average sampling time of about 31 seconds for a single image of $1024\times 1024$. 
As a training-free model, FreeStyle inevitably requires appropriate adjustment of hyperparameters to balance the intensity of style and content. 
In our qualitative experiments, we set the hyperparameters with $n=160$, $\sigma=958$, $b\in\left(0.5,3\right)$, and $s\in\left(0.5,2.5\right)$. We use the DDIM sampler to execute a total of 30 sampling steps for each image generation. Our model is based on the SDXL~\cite{sdxl}, utilizing its publicly available pre-trained model as the model parameters for inference. 

\begin{figure*}[!t]
\centering
\includegraphics[width=\textwidth]{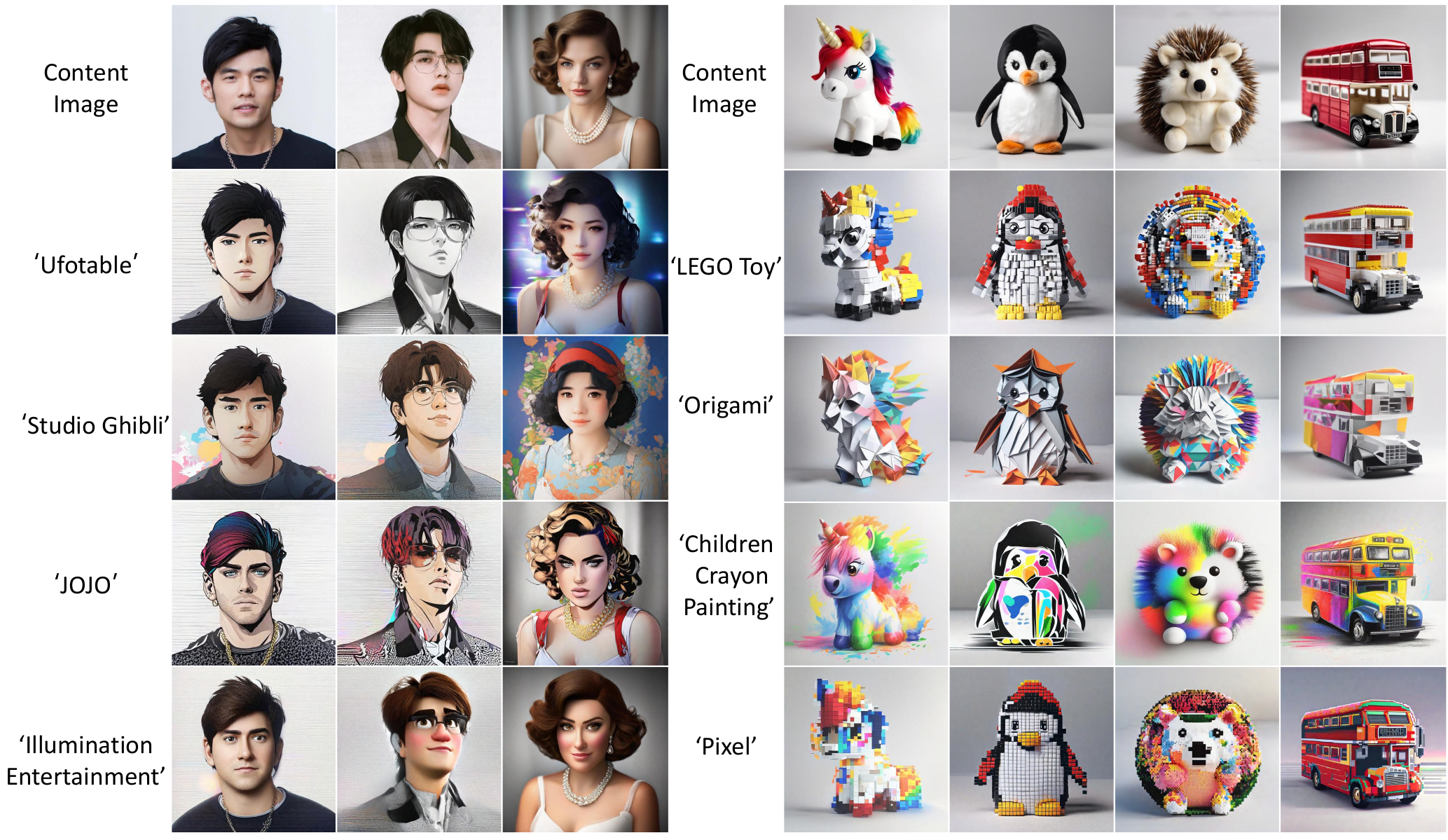}
\caption{The results of style transfer on portraits (left) and objects(right) using FreeStyle. Under the conditions of fine-grained anime style (left) and physical style(right) text prompts, the stylized results achieved with FreeStyle still exhibit clear fine-grained style differences and achieve a natural fusion of style and content.}
\label{result2}
\end{figure*}

\begin{figure*}[ht]
\centering
\includegraphics[width=\textwidth]{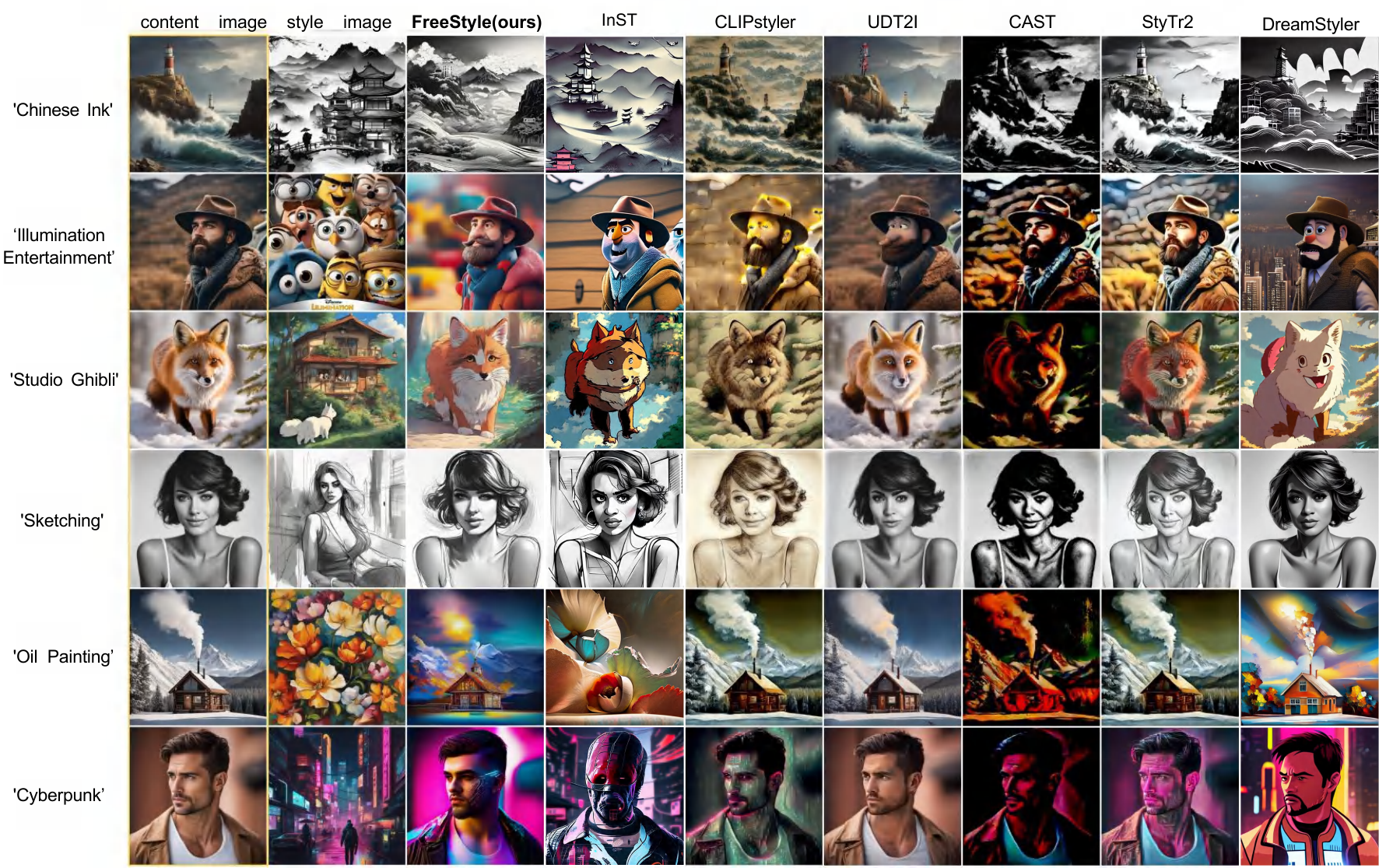}
\caption{Qualitative comparison with several state-of-the-art image style transfer methods, e.g., InST~\cite{InST}, CLIPstyler~\protect\cite{clipstyler}, UDT2I~\protect\cite{uncoverdis}, CAST~\protect\cite{CAST}, StyTr$^2$~\protect\cite{stytr2} and DreamStyler~\cite{dreamstyler}.}
\label{contractive}
\end{figure*}

\subsection{Experimental Result}
\noindent\textbf{Qualitative Results.} 
To verify the robustness and generalization ability of FreeStyle, we conduct numerous style transfer experiments with various styles across different content. 
Fig.~\ref{result1} presents the style transfer effects of FreeStyle in the domains of buildings, landscapes, animals, etc. 
The experiments include style transfer in ``Chinese Ink'', ``Embroidery Art'', ``Oil Painting'', ``Watercolor Painting'', ``Studio Ghibli'', ``Cyberpunk'', ``Pixel Punk'', ``Wasteland'' and ``Sketching''. 
We showcase the results of applying style transfer to human portraits using FreeStyle, as in Fig.~\ref{result2} (left). 
In this figure, we also conduct style transfer experiments with multiple styles, including ``Ufotable'', ``Studio Ghibli'', ``JOJO'' and ``Illumination Entertainment''. 
Observations indicate that FreeStyle is capable of providing accurate style information for the style transfer results while almost completely preserving the content information. 
For instance, the stylized results for ``JOJO'' maintain the structural information, while reasonably adjusting the image according to the character traits in the ``JOJO'' anime, like bold outlines, strong lines, and vibrant coloring. This achieves a more natural fusion and expression of both style and content.
It is noteworthy that we perform style transfer on images using fine-grained styles from four animation categories in Fig.~\ref{result2} (left). Despite this, FreeStyle is still able to achieve style transfer results with high recognizability and accurate styling. Additionally, we apply multiple physical style transfers to various everyday items. As illustrated in Fig.~\ref{result2} (right), FreeStyle demonstrates excellent style transfer effects across these styles.

\noindent \textbf{Qualitative Comparisons.} 
As shown in Fig.~\ref{contractive}, we conduct extensive comparative experiments with state-of-the-art methods, covering various styles and diverse content images. 
The results show the apparent advantages of our method over others, as it can reasonably modify shapes (e.g., rows 1,2,6), brushstrokes (e.g., rows 1-5), lines (e.g., rows 3,4), and colors (e.g., rows 1-6) to achieve superior artistic effects. 
In comparisons between our method and several others, it is noticeable that our approach more accurately achieves style expression (e.g., rows 2,3,6), especially in styles that are more challenging to transfer. 
In the results of the 5th line for both InST~\cite{InST} and DreamStyler~\cite{dreamstyler}, varying degrees of leakage issues were observed. In contrast, FreeStyle avoids such problems by not using style images for the injection of style information. 
Additionally, compared to our method, style transfer results of InST~\cite{InST} excessively and unnecessarily alter the content information. 
A key objective of style transfer tasks is to adapt to the target style while preserving the integrity of the content information as much as possible. 
Results from CAST~\cite{CAST} and StyTr$^2$~\cite{stytr2} are often marked by noticeable halo effects (e.g., rows 3,5,6) and are blurred (e.g., rows 2,6). 
In contrast, FreeStyle can produce clear stylized images without any noticeable halo effects. 
The transfer results of both CLIPstyler~\cite{clipstyler} and UDT2I~\cite{uncoverdis} exhibit issues of failed and inaccurate style expression. 
In summary, Fig.~\ref{contractive} indicates that our method exhibits greater robustness, more accurate style expression, and more artistic style transfer effects. 

\noindent \textbf{Quantitative Comparisons.} 
To better evaluate our method, we employed multiple quantitative metrics for assessment, the results of which are presented in Tab.~\ref{table1}. 
For all comparison methods, we utilized their publicly available pretrained parameters for sampling. 
Following the widely used quantitative experimental setup~\cite{InST,clipstyler}, we performed style transfers on 202 content images including landscapes, architecture, people, and animals, across 10 styles (``Chinese Ink'', ``Illumination Entertainment'', ``Embroidery Art'', ``Graffiti Art'', ``Impressionism'', ``Oil Painting'', ``Watercolor Painting'', ``Cyberpunk'', ``Studio Ghibli'', ``Sketching''), resulting in a total of 2020 stylized images for each method. For the CLIP Score~\cite{CLIP}, we calculate the cosine similarity between the CLIP image embeddings and the prompt text embeddings. Using the prompt as a style description, we believe that a higher CLIP Score indicates a more accurate expression of style. 
The CLIP Aesthetic Score evaluates the quality, aesthetics, and artistic nature of images using a publicly available pre-trained art scoring model. A higher CLIP Aesthetic Score indicates that the fusion of style and content is more aesthetically pleasing. 
Training Cost refers to the product of the number of parameters that need to be optimized during the training phase and the recommended number of iterations in the corresponding method. 
FreeStyle achieved state-of-the-art (SOTA) results in both the CLIP Aesthetic Score and Training Cost, as shown in Tab.~\ref{table1}. Additionally, FreeStyle demonstrated competitive results in the CLIP Score.


\begin{table}[t]
\centering
\renewcommand{\arraystretch}{1.23}
\caption{Quantitative comparisons with state-of-the-art methods are conducted, using CLIP Aesthetic Score, CLIP Score and Training Cost as our evaluation criteria. }
\resizebox{0.7\textwidth}{!}{
\begin{tabular}{cccc}
\specialrule{0.1em}{0pt}{0pt}
           & CLIP Aesthetic Score $\uparrow$ & CLIP Score $\uparrow$ & Training Cost$\left(\sim\right)$ $\downarrow$ \\ \hline
CAST~\cite{CAST}       & 5.1462               & 22.347      &3.51M$\times$400      \\  \hline
StyTr2~\cite{stytr2}     & 5.8613               & 22.300      & 35.39M$\times$0.16M       \\ \hline
CLIPstyler~\cite{clipstyler} & 6.0275               & \textbf{27.614}      & 0.62M$\times$200     \\ \hline
UDT2I~\cite{uncoverdis}      & \underline{6.2290}               & 21.708       & \underline{50$\times$ 10}     \\  \specialrule{0.05em}{0pt}{0pt} \rowcolor{gray!20}
\textbf{FreeStyle (ours)}  & \textbf{6.3148}               & \underline{25.615}          & \textbf{0M}    \\ 
\specialrule{0.1em}{0pt}{0pt}
\end{tabular}
}
\label{table1}
\end{table}


\subsection{Ablation Study}

\begin{figure}[h!]
\centering
\includegraphics[width=\textwidth]{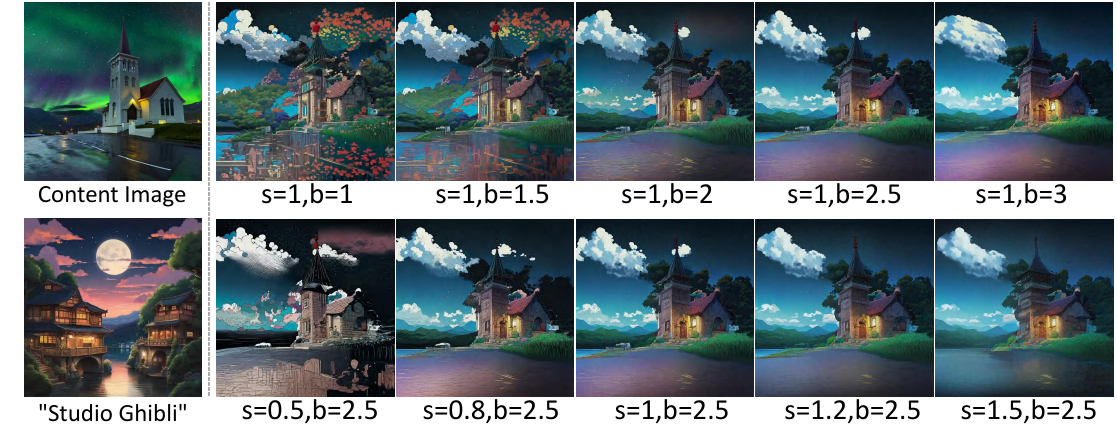}
\caption{The ablation study of hyper-parameter $s$ and $b$.}
\label{ablation0}
\end{figure}

\noindent \textbf{Effect of hyperparameters $b$ and $s$.} We present the results of ablation experiments conducted on hyperparameters $b$ 
\noindent and $s$, as in Fig.~\ref{ablation0}. 
In FreeStyle, the intensities of content and style information are adjusted by the two hyperparameters $b$ and $s$, respectively. 
From the experimental results, when $s$ is fixed and $b$ increases, the content information of the image becomes clearer and more complete. Conversely, when $b$ is fixed and $s$ increases, the style expression of the image becomes gradually more accurate and enhanced. 
However, there is also a relatively inhibitory relationship between them. When $s$ is fixed and $b$ decreases, the image shows more Ghibli-style clouds and plants. When $b$ is fixed and $s$ decreases, the results exhibit more restored content outlines. 
To clarify this feature more clearly, we will provide further explanation in the subsequent ablation experiment on $s$.

\begin{figure}[!ht]
\centering
\includegraphics[width=\textwidth]{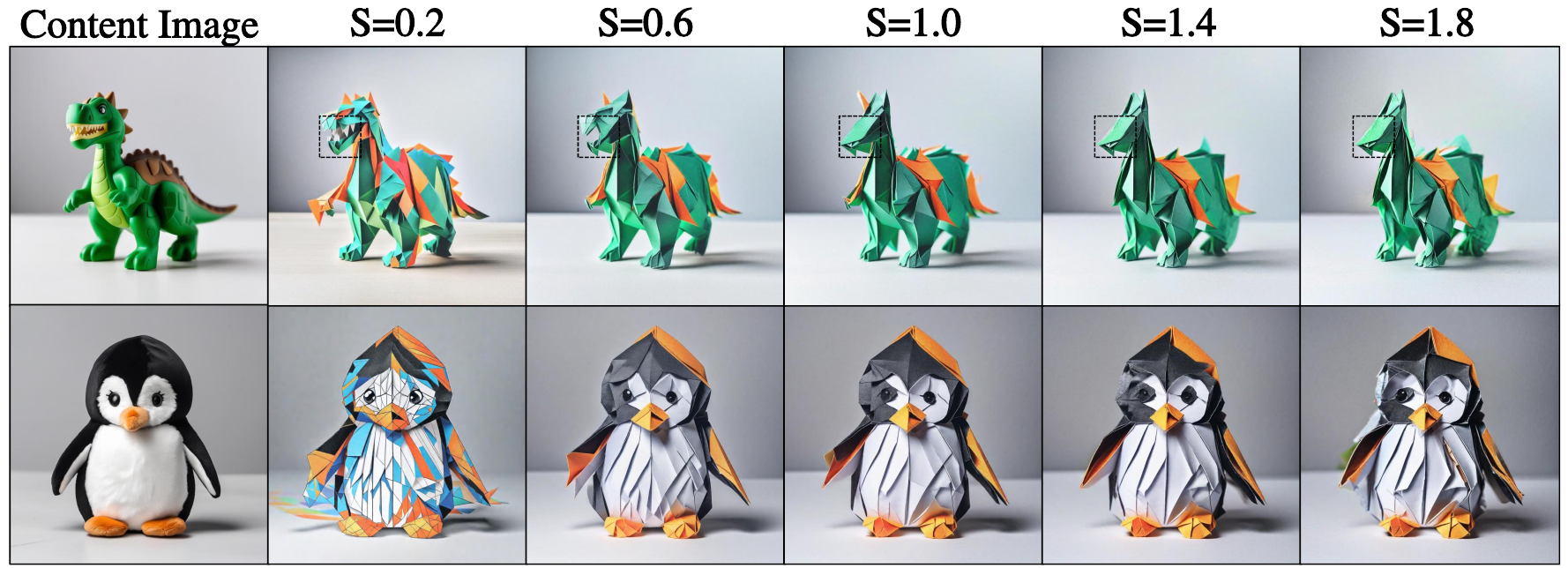}
\caption{Ablation experiment on the impact of the hyperparameter $s$ on style intensity.}
\label{abls}
\end{figure}

\noindent \textbf{Effect of hyperparameters $s$.}
We conduct ablation experiments on the transfer of ``origami art'' style using different settings of $s$, in Fig.~\ref{abls}. It is evident that adjusting the hyperparameter $s$ significantly affects the intensity of the style in the images. As $s$ increases, the style intensity enhances while the content information is relatively diminished. Conversely, reducing $s$ weakens the style intensity and can even lead to inaccuracies in style expression, as seen in the second row of Fig.~\ref{abls} where $s=0.2$.

\begin{figure}[!ht]
\centering
\includegraphics[width=\textwidth]{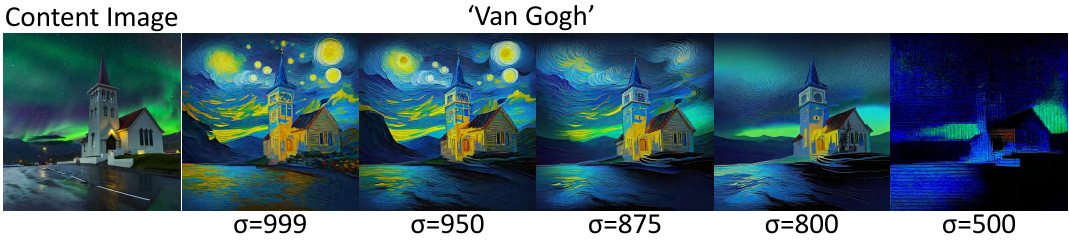}
\caption{The ablation study of hyper-parameter $\sigma$.}
\label{ablation1}
\end{figure}

\noindent \textbf{Effect of hyperparameter $\sigma$.} Fig.~\ref{ablation1} illustrates the impact of the hyperparameter $\sigma$ on the style transfer effect. The observations indicate that better style transfer are achieved when $\sigma$ exceeds 850, whereas the effect gradually deteriorates as $\sigma$ becomes too small. 
We believe a too small $\sigma$ value makes $f_s$ contain excessive content information, significantly disrupting the style information. 
However, in our experiments, we find that setting the parameter $\sigma$ to 958 and not requiring manual tuning resulted in good performance across all images. 

\begin{figure}[!ht]
\centering
\includegraphics[width=\textwidth]{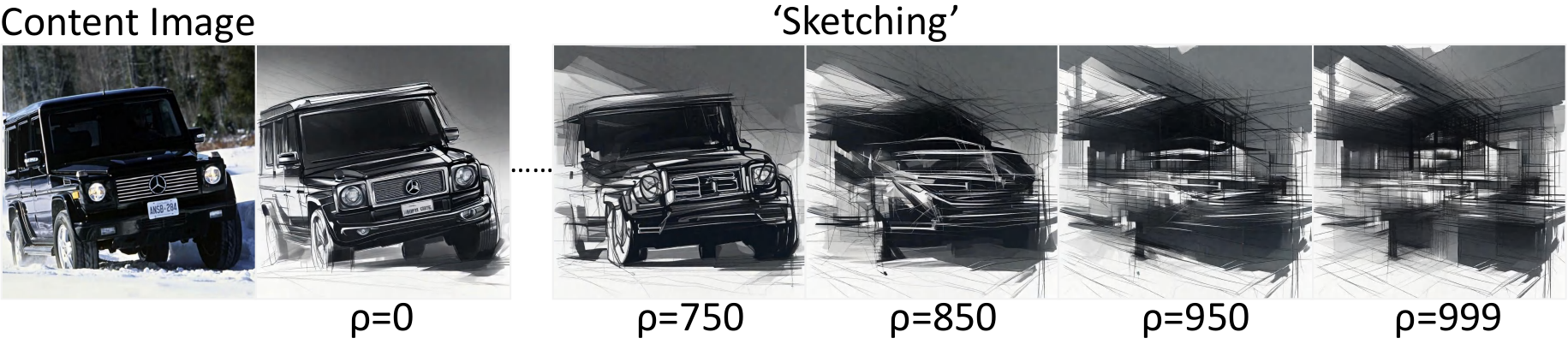}
\caption{An ablation study where varying levels of noise are added to the content image input $x_0$ to eliminate content information. (the larger $\rho$, the more noise is introduced)}
\label{jeepab}
\end{figure}

\noindent \textbf{Content-Style Disentanglement.} To further validate FreeStyle's ability to disentangle content and style information, we introduced varying degrees of $\rho$ noise into the input $x_0$ of the content feature $f_c$ to reduce content information input and observed the preservation of content and style information. 
As shown in Fig.~\ref{jeepab}, with the increase of $\rho$ and hence more noise, the content information in $f_c$ gradually decreases while the style feature $f_s$ remains unchanged. 
It is clearly observed that as the value of $\rho$ increases, content information progressively decreases without affecting the expression of style information. When $\rho = 999$, content information almost completely disappears, yet the expression of ``sketching'' style lines and brushstrokes remains observable. This validates FreeStyle's powerful capability in disentangling content and style information. 

\begin{figure}[!h]
\centering
\includegraphics[width=\textwidth]{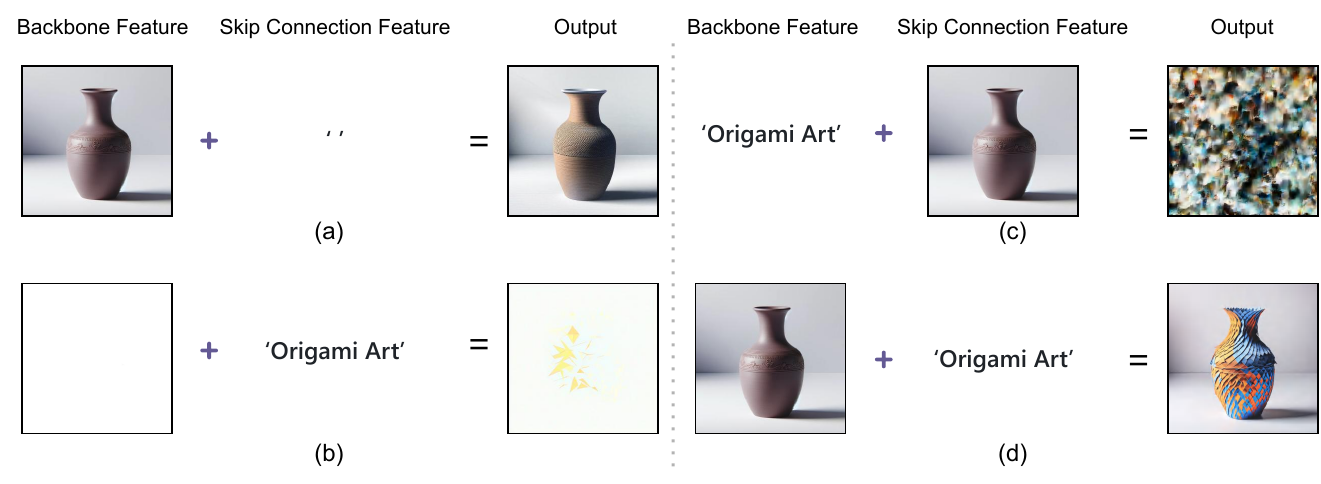}
\caption{The ablation study evaluates the decoupling ability of U-Net.}
\label{decouplingstudy}
\end{figure}

\begin{figure*}[!t]
\centering
\includegraphics[width=\textwidth]{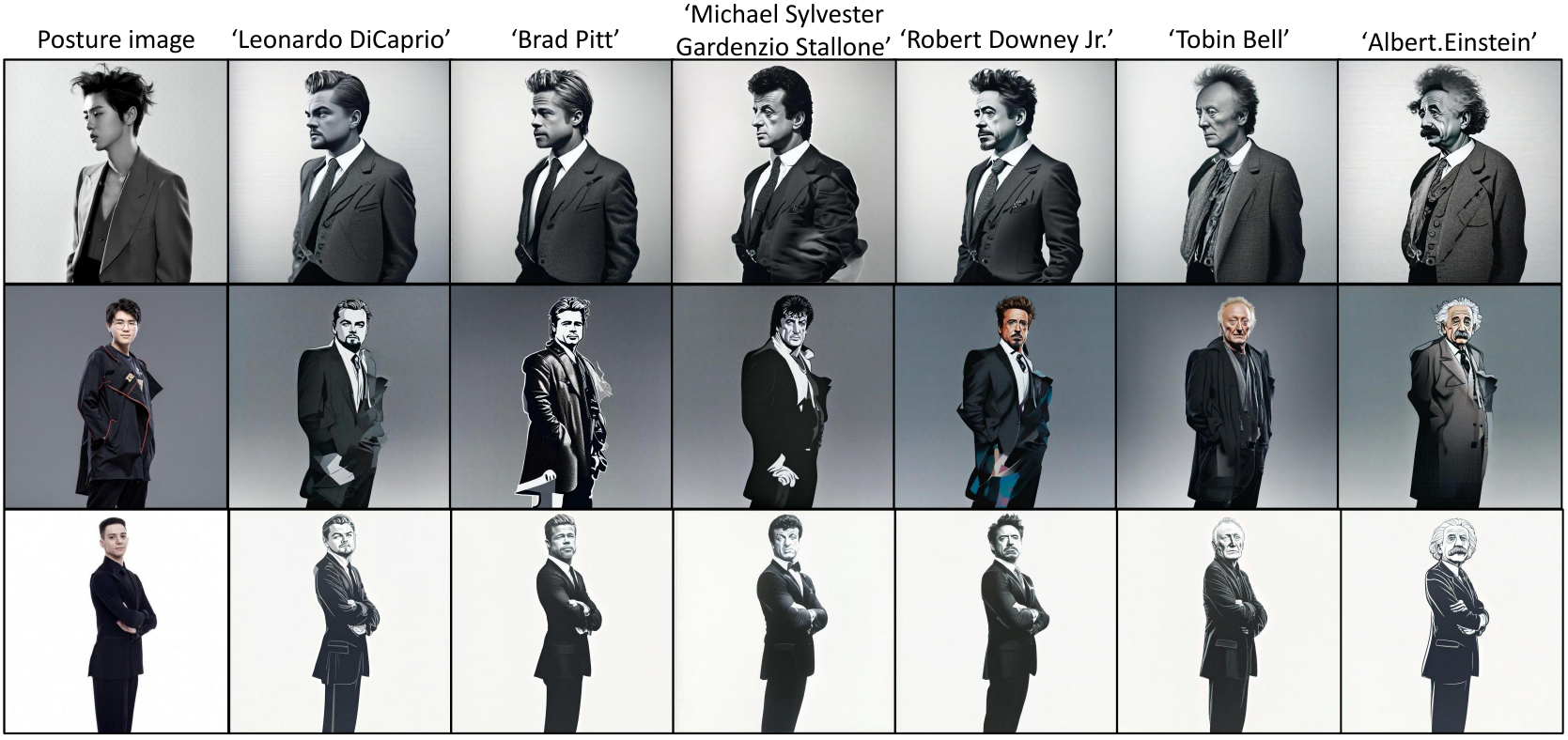}
\caption{In FreeStyle, when replacing the content image with a posture image and using a person's name as the prompt, it is clearly observable that the characters in the generated images maintain the same pose.}
\label{peter}
\end{figure*}

\noindent \textbf{Ablation of U-Net Decoupling} 
To further verify that the U-Net structure has the ability to decouple content and style, we conducte the ablation study shown in Fig.~\ref{decouplingstudy}. 
Specifically, we conducte experiments by controlling variables as follows: (a) the backbone network input features are content features, and the skip connection features are replaced with null features, (b) the backbone network input features are replaced with null features, and the skip connection features are style features, (c) the backbone network input features are style features, and the skip connection features are content features, (d) the backbone network input features are content features, and the skip connection features are style features. 
From the experiments in Fig.~\ref{decouplingstudy} (a), (b), and (d), we can easily demonstrate that the model can control the output of image content and style information through the backbone features and skip connection features, respectively. In Fig.~\ref{decouplingstudy} (a), when the skip connection input is null, the model output still maintains the image content information unchanged. In Fig.~\ref{decouplingstudy} (b), when the backbone feature input is null, the model still preserves the style in the skip connection features. 
In Fig.~\ref{decouplingstudy} (c) and (d), we swapped the backbone features and skip connection features and observed the model's generated results. We found that when content feature is used as the skip connection feature and style feature is used as the backbone feature input to the network, the model fails to generate effective results. This further verifies that the backbone features and skip connection features of U-Net correspond to the content and style information of the generated images, respectively.

\subsection{Other Study}
Based on the exploration and analysis of the U-Net structure in this paper, we believe that the backbone network's ability to suppress high-frequency information and the predominance of high-frequency information in skip connection features can be used to achieve a variety of other interesting effects. Fig.~\ref{peter} illustrates how we used a posture image to achieve uniform pose generation. 
Following this, we replace the content image with a posture image as the image input, and a person's name replaces the style input as the prompt input. 
We observe that each row in the figure has generated characters consistent with the prompt, and these characters maintain the same pose as in the posture image. 
Since we did not include any style information in the prompt, which controls the generation of style features, we observe that the generated style exhibits noticeable inconsistency and uncertainty. In contrast, the consistency of the characters' poses in the image is due to using the pose image as the content image input, thereby preserving the overall structural information of the image. These results indicate that this method has the potential to achieve specific pose generation after fine-tuning on a particular object. 

\section{Conclusion}

In this study, we present FreeStyle, an innovative text-guided style transfer method that utilizes pre-trained large text-guided diffusion models. Diverging from previous approaches, FreeStyle accomplishes style transfer without the need for additional optimization or reference style images. The framework, comprising a dual-stream encoder and a single-stream decoder, seamlessly adapts to specific style transfer tasks by adjusting scaling factors. Despite its simplicity, our method showcases superior performance in terms of visual quality, artistic consistency, and robust preservation of content information across diverse styles and content images. These findings significantly advance the field of training-free style transfer. 
Meanwhile, as a training-free approach, the unavoidable manual parameter tuning remains an area for improvement. 
In future work, we will address this issue to achieve a parameter-adaptive training-free style transfer method.

\bibliographystyle{elsarticle-num} 
\bibliography{sn-bibliography}



\end{document}